\documentclass[12pt]{article}

\usepackage{PRIMEarxiv}

\usepackage[utf8]{inputenc} 
\usepackage[T1]{fontenc}    
\usepackage{hyperref}       
\usepackage{url}            
\usepackage{booktabs}       
\usepackage{amsfonts}       
\usepackage{nicefrac}       
\usepackage{microtype}      
\usepackage{fancyhdr}       
\usepackage{graphicx}       
\usepackage{tabularx}       
\usepackage{listings}       
\usepackage{xcolor}

\graphicspath{{media/}}     

\title{Traccia: 

An OpenTelemetry-Based Governance Platform for AI Systems}

\author{
  Nutan Kumar Naik \\
  Department of Earth and Atmospheric Sciences \\
  National Institute of Technology Rourkela \\
  Odisha, India \\
  \texttt{521er1007@nitrkl.ac.in} \\
  \And
  Nutan Kumar Naik, Aditya Kumar Saroj, Vijay Prasad Poudel, Saurav Samantray, Abhishek Patel \\
  Algen.AI \\
  Bengaluru, Karnataka, India \\
  \texttt{nutan@algen.ai}, \texttt{aditya@algen.ai}, \texttt{vijay@algen.ai}, \texttt{saurav@algen.ai}, \texttt{abhi@algen.ai} \\
}

\begin{document}
\maketitle

\begin{abstract}
The rapid development of Large Language Models (LLMs) and Artificial Intelligent (AI) powered autonomous agents has fundamentally changed the existing forms of software governance. In spite of the rigorous standards of transparency and accountability required according to the international frameworks such as the European Union’s AI Act, there is a considerable gap between theory and reality. The present study discusses the inherent drawbacks of currently utilized platforms for LLM evaluation, machine learning workflow, and application performance monitoring in general. It has been shown that current disjointed solutions fail to protect unbound state space agentic architecture from serious threats such as alignment drift, SaaS security concerns, and unauthorized deployment of shadow AI systems. Moreover, a solution is proposed for overcoming the discussed challenges in form of a coherent multi-level AI governance stack Traccia built on the top of OpenTelemetry infrastructure platform. Traccia resolves the last mile for AI Alignment by adding the telemetry data, passive semantic guardrail assessment, and execution lineage into a hashed trace ledger. Traccia automatically creates compliance evidence packages by appending tamper-resistant fingerprints and SHA-256 content hash, that map to regulatory requirements (Articles 12, 14, 19, 26(6), and 50 of the EU AI Act) without invading any data privacy. By performing this evaluation in a methodical manner, a solid machine-readable base has been created for enterprise-wide management of autonomous AI systems.
\end{abstract}


\section{Introduction}

Integration of Artificial Intelligence (AI) has led to a paradigm shift in software engineering from deterministic, rules-based workflow to a probabilistic, highly adaptive and distributed AI architectures \cite{sapkota2025ai, tsuei2026ai}. The core of this change is due to the use of Large Language Models (LLMs). These complex models work not only as simple text generators but as independent agents in a socio-technical environment \cite{yang2025from, yao2025a}. Through the use of Retrieval Augmented Generation (RAG), recursive tool invocation and multi-agent coordination, these systems perform open-ended tasks in huge, non-deterministic spaces \cite{yang2025from}. Such an approach creates instability in traditional assumptions of software control, data provenance and predictability \cite{patra2026safetyconstrained, sengupta2025interpretability}. This brings unpredictable system dynamics, decision loops that act independently, and execution problems that regular software assurance cannot handle. Quality assurance of traditional software includes static analysis, unit tests, and benchmarking in bounded conditions \cite{hagendorff2020the, tabassi2023artificial}. Ex-ante evaluation does not consider failure modes emerging in dynamic environments that have unbounded states of AI-based systems \cite{chowdhury2022unlocking, chua2026shadow, wass2026global}. After deployment, agentic systems can suffer from alignment drift, contagion across agents, prompt injection, and adversarial exploitation \cite{collins2026systemic, westover2026when}. Assuming adaptive AI agents as static state machines is an essential oversight that poses a significant risk since offline assessments do not help detect the deviation from the intended behaviour \cite{demirhan2026llmbased, nguyen2025probing}.

In such dynamic environments runtime observability is essential to establish a firm foundation for AI systems. Instead relying on predictions of state transitions, the architectures now have to gather runtime telemetry data such as logs, performance metrics, and distributed traces \cite{bandara2026ai, morgan2026ai}. On systems levels, anomaly detection can be done by monitoring the signatures of GPU usage, memory usage, and system-level calls, which incur with minimal overhead \cite{rahman2026detecting, spadacini2026clinical}. On the application level, distributed tracing keeps track of the execution of queries made by the LLMs, their retrieval process, and the interaction between external tools \cite{vigeant2026a}. More importantly, there is a clear distinction that needs to be drawn between observability and governance at runtime. Observability is descriptive, while governance is prescriptive, i. e., logging information about the execution process does not actually enforce policies or stop undesirable behaviour of models \cite{meyman2026observability}. Raw telemetry in enterprise scenarios does not have structure and durability; it is not cryptographically verified and not consistent with high-level policies \cite{ugarte2026making}. Governance involves a translation of the raw telemetry to structured machine-verifiable operation evidence that is consistent with safety envelopes in real-time \cite{bandara2026ai}. Otherwise, an integration of the telemetry with enforcement engines is impossible, and the observability dashboard does not function as a compliance construct.

There have been significant studies on AI governance taxonomies \cite{sengupta2025interpretability}, standards for tracing \cite{yu2026backcasting}, and monitoring architectures \cite{bandara2026ai}; however, a crucial gap remains among them. In particular, relatively less effort has given to the analysis of how execution telemetry can be transitioned into structured governance evidence. The current observability techniques are lacking the semantic understanding required for rapid injections, agentic tool commands, and eXplainable AI (XAI) measures \cite{seth2025bridging}. Moreover, there is no clear algorithm showing how to use tracing primitives to collect cryptographic evidence of compliance \cite{morgan2026ai}. To fill this gap, this study provides a systematic architectural analysis of Traccia (\url{https://traccia.ai/}), which is based on OTel standards and functions as an AI governance framework. Traccia works directly during the execution path, governs foundation models or agents, by extending the standard OTel semantic conventions. Through such architecture, Traccia systematically translates low-level signals, such as prompts, tokens, tool calls, and safety triggers, into structured and policy-compliant compliance artifacts. The rest of the article is organized as follows: section 2 provides a summary of the fundamentals of AI governance, observability, and current governance platforms, whereas section 3 discusses various evolving AI governance architectures and outlines the need for continuous runtime compliance. Finally, Traccia is introduced in section 4 as a means to convert runtime telemetry into governance and compliance artifacts via its architecture, and the final Section 5 draws summary and future works.

\section{Background and Related Works}

\subsection{AI Governance Taxonomy}

The complexity of modern AI architecture makes it impossible to consider its governance using a unified approach. Reduction of AI into a single piece of software is bound to lead to an inevitable obscuring of different epistemological and operational boundaries within the AI workflow \cite{chowdhury2022unlocking, chung2024false, siala2022shifting, tripathi2026mapping, varnosfaderani2024the}. Therefore, implementation of a taxonomy-based approach to decompose AI architectures in order to properly match regulatory requirements on each of the structural level is essential \cite{deshazer2026the, morley2022governing}. Following this principle, the section outlines a conceptual framework for AI governance structured into four main classes: model-level, pipeline-level, runtime, and compliance governance, which together cover the whole spectrum of potential risk mitigation techniques \cite{chowdhury2022unlocking, ferrara2026defeat, sumner2026the}.

Model-level governance relates to the inherent attributes of the various strategies followed during  machine learning model development, which includes training processes, data origins, and alignment methods. At this bottom-most level, the emphasis in governance strategies lies in the interpretability of mechanisms, latent space regularity, and imposition of behavioural priors through the approaches such as Constitutional AI or Reinforcement Learning from Human Feedback (RLHF) \cite{gkta2025shaping, idowu2026mapping, patra2026safetyconstrained, sunu2026persistent, yang2025from, yu2026backcasting}. Also, it is acknowledged that while strategies such as red-teaming and safety tests pre-deployment may serve to approximate governance framework alignment, they continue to be limited within the bounds of the models \cite{nguyen2025probing, ugarte2026making, westover2026when}. Therefore, it becomes imperative that governance at the model level takes into consideration the audibility of the training corpus to prevent the systematic embedding of biases, copyright infringements, and dangerous concepts within the system \cite{kim2026people, unver2024ethics}. Moving from the mathematical abstraction used in the theory to software engineering implementation, pipeline-level governance impose controls the Machine Learning Operations (MLOps) workflow. It regulates the reproducibility of experimental ML analysis, the cryptographic embedding model weights, and keeping track of hyperparameter configurations \cite{cho2026large, hrer2025specification, nguyen2025probing}. The most agreed upon idea among software assurance experts is the necessity of a broader Software Bill of Materials (SBOM) for AI. It helps track dependency history, third-party libraries, and dataset hashes \cite{rahman2026detecting, spadacini2026clinical}. By standardizing the Continuous Improvement/Continuous Development (CI/CD), pipeline management eliminates supply chain risks and ensures that the model which gets to production is computationally identical to the model audited ex-ante \cite{sumner2026the, williamson2024balancing}.

Runtime governance represents the defining edge of AI observability, which focuses on the emergent and hard-to-predict behaviour of the models in response to adversaries or novel input distribution structures in production operations. In contrast to deterministic applications, LLMs and agentic AI must have continuous zero-trust inference monitoring \cite{mello2026the, sharma2025maltopic, yao2025a}. It relies mainly on telemetry and includes various anomalies, and semantic inconsistencies using execution traces \cite{rahman2026detecting, spadacini2026clinical}. The literature comprehensively covers the trend towards guardrails detection that intercept and analyse dynamic tool usage, and inter-agent communication in real-time. It also enforces threshold-based circuit breakers that segregate rogue agents prior to causing system-wide breakdowns due to hallucinations. Compliance governance is at the top of the hierarchy that acts as a transition layer connecting the technical execution telemetry with legal regulatory frameworks \cite{tian2026from}. The main goal of this layer is to produce machine-readable, legally binding evidence in conformity to systemic risk requirements. This layer uses the information from the previous layers, (e. g., model, pipeline, and runtime) to create standard embedded ledgers or transparency registers \cite{hrer2025specification, mkander2023auditing, seth2025bridging}. Researchers argue that a strong compliance governance is impossible without the unbreakable chain of custody that immutably connects high-level policy statements to the low-level observations of the deployed infrastructure \cite{qian2026large, tripathi2026mapping, zahid2026public, zhang2019artificial}.

The integrity of the AI safety structure strongly depends on a bidirectional flow of information between the above-listed four layers. The recurring issue mentioned in the literature is the last mile challenge, where governance policies and compliance requirements fail to translate into verified runtime controls and evidence \cite{adapa2026navigating, bateyko2025one, ingrams2022transparencys}. The lack of connection between high-level policy and low-level operation telemetry often turns compliance into largely document-driven rather than evidence-based. In contrast, telemetry in the absence of any meaningful semantics produces an enormous amount of noise, which are difficult to interpret for governance and audit \cite{rahman2026detecting, spadacini2026clinical}. These observations together indicate that establishing traceable semantic links between governance policies, execution events, and runtime evidence is an essential requirement for achieving continuous and verifiable AI governance \cite{patra2026safetyconstrained, sharma2025maltopic}.

\subsection{Observability Foundations for AI Systems}

The implementation of the multi-layered governance framework described in the previous section relies entirely upon the shift from software-based monitoring to systemic observability. While the former simply tells us whether a system is working, the later analyses the inner events of complex systems based on their output \cite{danks2026development, killoran2026learn}. An extensive study of the existing literature shows that various AI models and agents cannot be monitored using conventional black-box approaches due to their non-deterministic output and altering state spaces \cite{rahman2026detecting, spadacini2026clinical}. As such, the development of observability primitives at the software level is seen as an absolute necessity \cite{dadzie2025multilayered, hatz2025the}.

The observability of distributed systems is founded on three basic principles: logs, metrics, and traces. However, observing AI systems requires a considerable semantic rethinking of these basic principles \cite{rahman2026detecting, spadacini2026clinical}. In terms of LLMs, logs are no longer simple error logs but rather multi-modal semantic events such as prompt injection, safety filter, and unfiltered output from the model itself \cite{hazime2025evaluation, mkander2023auditing}. In addition, metrics no longer involve CPU utilization and network latencies but rather represent continuous algorithmic behaviours like token production efficiency, inference computation (Floating Point Operations Per Second), distribution drift, and inference cost in real time \cite{rahman2026detecting, spadacini2026clinical}. Most importantly, traces serve as the crucial connectors that link separate logs and metrics in order to construct a chronologically ordered narrative on reasoning steps taken, Application Programming Interface (API) requests made, and memory lookups performed by AI agents \cite{killoran2026learn, whittlestone2022ai}. In comparison to the other two pillars (logs and metrics), distributed tracing becomes the key architecture of AI governance. Moreover, AI architectures are typically distributed, not a monolith and rely on vector databases, different tools, and orchestration microservices \cite{mello2026the, odubola2024ai}. The distributed tracing architecture works by modelling execution as a Directed Acyclic Graph (DAG), in which each execution has a set of hierarchical ‘spans’ associated with it. These spans define different units of work being performed, which can include various events, generating a completion, among other attributes, their latencies, and execution lineage \cite{rahman2026detecting, spadacini2026clinical}. Trace tree is used to model AI governance architecture to ensure that the source of all forms of hallucinations or policy violations can be easily traced \cite{cho2026large, mkander2023auditing, nguyen2025probing}.

The need for distributed tracing reaches its peak within the domain of cutting-edge LLM architecture and agentic loops. Contrary to standard machine learning inference, which uses one call to model’s weights, the modern AI agents work within frameworks, such as LangChain or AutoGen, which enable recursion through reflection, dynamic tool creation, and inter-agent discussions \cite{gkta2025shaping, yu2026backcasting}. The tracing of such a system needs to include not only the end result of the agent’s actions, but also the chain of thought, the exact contextual information obtained using RAG, and the changes within the agent’s systemic state due to execution of external code \cite{spadacini2026clinical}. However, the lack of such detailed trace makes it impossible to audit the agent’s compliance after deployment because the reasoning behind the action becomes impossible to retrieve \cite{mkander2023auditing, seth2025bridging}. In order for the tracing process to bridge the gap between technical implementation and compliance regulations, the telemetry must be standardized. Since various orchestration frameworks produce completely heterogeneous logs, it is difficult to aggregate evidence into a consistent transparency register \cite{bengio2024managing, yu2023leveraging, zhang2019artificial}. In order to address this problem, a number of semantic conventions related to the AI technology have been established, such as standardized attribute schemas (e. g., llm.prompt, llm.completion\_tokens, llm.system, etc.) that regulate the way in which interactions within the AI system should be captured in the trace records \cite{binkyte2025interactional, kim2025ai}. Through the application of such strict ontological schemas in the complete AI lifespan, enterprises ensure that traces stay readable and interpretable. This point is repeatedly emphasized in scientific works as an essential requirement for conducting automated audits using various governance framework \cite{rahman2026detecting}.

In conclusion, observability is not just an operational process in the domain of site reliability engineering but the establishes the foundation for AI governance. Also, it is demonstrated that regulations are not enforceable in the absence of proof that can be provided only through continuous telemetry \cite{akgn2026governing, barak2026the, dharmansyah2026privacy}. Spans, traces, and token metrics make up the empirical data that is used in the compliance governance layer. The transformation of opaque and non-deterministic processes of AI systems into chronologically structured and cryptographic registers is what observability does to provide immutable ground truth \cite{ferrara2026defeat, sumner2026the}.

\subsection{Existing AI Governance and Observability Tools}

The implementation of the above-mentioned taxonomy of governance practices has led to a development of numerous specialized software applications. Moreover, the market for safety and observability of AI is rather fragmented and evolving simultaneously within different spheres of engineering \cite{danks2026development, ramcharran2025orchestrating, yu2023leveraging}. Three major types of tools are highlighted in the literature: the evaluation platforms concentrated on the semantic behaviour of AI systems, the experiment tracking applications based on the classic MLOps workflow, and the observability systems generalized from cloud-native microservices \cite{demirhan2026llmbased, hrer2025specification, minkkinen2022what, mkander2023auditing}. In this section, the current paradigms of these tools are critically examined in order to illustrate both their usefulness and insufficiency in meeting the regulatory requirements.

	The first set of platforms refers to the tools that are explicitly built to regulate the semantics and behaviour of LLMs. Such platforms like TruEra, Arize AI (Phoenix), LangFuse, and LangSmith are highly inclined towards prompt assessment, RAG scores, and hallucination metrics \cite{korinek2022aligned, odubola2024ai, sengupta2025interpretability, xie2026ethical}. The aforementioned platforms are good at deploying heuristic guardrails that include detection of toxicity, Personally Identifiable Information (PII) leaks, and prompt injections at the software API boundary \cite{hrer2025specification, nejadgholi2025a, nguyen2025probing}. With the help of ‘LLM-as-a-Judge’ and distance measures, these platforms offer much-needed semantic visibility \cite{rahman2026detecting, spadacini2026clinical}. The second category belongs to the traditional MLOps methodology. MLflow, Weights \& Biases (W\&B), and CometML are typically built with the intention of monitoring hyperparameters and various model artifacts \cite{venugopal2026evaluating}. These platforms have also incorporated LLM registries, prompt versioning, and strict dataset tracking into their architecture \cite{killoran2026learn, qian2026large, tsuei2026ai}. These systems are inherently positioned at the pipeline-level governance tier, where supply chain security and cryptographic model lineage are provided. Although these systems are essential for confirming that the model is identical to its predecessor, they do not have any visibility regarding the real-time state changes in inference level in the real world \cite{bandara2026ai, truby2020governing}.

	The third set of observability systems includes generally applicable, cloud-native observability suites like Datadog, New Relic, and Honeycomb. Initially designed to monitor the transition from monolithic applications to microservices, they have now added AI telemetry integration into their systems \cite{rahman2026detecting, spadacini2026clinical}. The main strength of these systems is telemetry. They also use OTel standards to collect very detailed metrics on the GPU memory usage, token generation time delays, and network distribution bottlenecks. Based on the examined literature, these observability systems collect the execution evidence at the fundamental level for zero-trust architectures of AI agents \cite{mello2026the, vigeant2026a, wawer2025integrating, yu2026backcasting}.

	Even considering the advanced abilities of these tools, there are critical structural flaws due to their isolated nature. The semantic evaluation toolkits do not provide the cryptographic hardware assurances required to identify any underlying tampering or evasion attack on the AI systems \cite{hrer2025specification, nguyen2025probing}. The second problem is that the classical experiment trackers do not work in real-time inference space. Therefore, they cannot serve as chain breakers in case of any rogue agentic behaviours \cite{gkta2025shaping, xie2026ethical, yu2026backcasting}. Additionally, general observability tools record physical execution but lack the semantic insight to determine whether an increase in the usage of compute resources is a normal data fetch or a disastrous jailbreak \cite{rahman2026detecting, spadacini2026clinical}. Overall, based on the current literature, the implementation of legal requirements, such as the European Union (EU) AI Act, by means of a fragmented multi-vendor toolchain can result in inevitable last mile problems \cite{aggarwal2023ai, debelak2026ai, ruschemeier2025experimental}. Nonetheless, the scientific community’s opinion makes it abundantly clear that standalone tools cannot be sufficient for regulating such state-space unbound systems. In order to address the need for technical guarantee and compliance, the industry should adopt unified and trace-based governance approach \cite{chung2024false, huang2025innovative, yu2018building}. Thus, a unified control stack that can connect semantic interests, pipeline provenance, and physical execution telemetry in a single immutable log is required. Therefore, by overcoming the restriction on traceability as a mere debugging tool and making it a cryptographic layer can secure the integrity of an AI system during its entire lifespan \cite{chen2021evaluating, ugarte2026making, venugopal2026evaluating}.

\section{AI Governance Regulations and Compliance Frameworks}

\subsection{The Evolving Regulatory Landscape}

With the fast adoption of AI, there is an increased move away from voluntary soft ethical principles to mandatory legal, technical, and organizational policies. The international community adopted high-level AI governance principles such as Organisation for Economic Co-operation and Development AI Principles and United Nations Educational, Scientific and Cultural Organization Recommendation on the Ethics of AI, which emphasizes abstract moral values like fairness, accountability, and transparency \cite{batool2025ai, seth2025bridging}. However, with the adoption of non-deterministic systems, limitations of purely ethical principles became obvious. Therefore, a more mandatory approach to regulation has emerged. For instance, in order to regulate the adoption of AI and agentic systems, European Union has recently introduced an AI Act \cite{kim2025ai}. This type of horizontal regulation takes place within a wider, multi-layered digital policy ecosystem, which includes General Data Protection Regulation (GDPR), Network and Information Systems Directive 2 (NIS2), and the Cyber Resilience Act (CRA) \cite{fabiano2026the}. Within this complex environment, corporate compliance requires proof of model behaviour which necessitates new professional positions such as AI Legal Specialist \cite{fabiano2026the}.

\subsection{Synthesizing Major Regulations and Standards}

Unlike isolated processes, global regulations and international standards create a multi-level compliance ecosystem with a set of common goals. According to the EU AI Act, high-risk AI systems have to meet strict lifespan requirements (Article 9-15) that include risk management, data governance, technical documentation, logging, transparency, human oversight, and operational accuracy \cite{jarvers2026engaged, ugarte2026making}. To fulfil these requirements of the legislation, companies align their processes in a framework. Non-sectoral and voluntary National Institute of Standards and Technology (NIST) AI Risk Management Framework (AI RMF) is aimed at helping employees recognize and mitigate risks through four major functions: Govern, Map, Measure, and Manage \cite{morgan2026ai}. At the same time, standards organizations have also created process-oriented specifications for management of AI systems, such as International Organization for Standardization (ISO) 42001:2023, which specifies an AI Management System (AIMS), and ISO 23894 that contains AI risk management guidelines \cite{fabiano2026the}.

AI-compliant frameworks must be integrated into existing enterprise-level security and privacy frameworks. This is because data protection laws such as GDPR and Health Insurance Portability and Accountability Act (HIPAA) contain provisions related to logging and explainable automation of decisions \cite{bandara2026ai, fabiano2026the}. Conventional security frameworks such as System and Organization Controls 2 (SOC 2) and ISO 27001, for instance, contain provision related to the systematic approach towards security and risk \cite{bandara2026ai}. Traceability, transparency, and accountability are key components of both AI and conventional security frameworks even though they are different in nature. ISO 42001 and SOC 2 define organization-level processes but do not offer a way to obtain technical and product-level evidence \cite{jarvers2026engaged}. In addition, existing compliance frameworks have been developed for static and deterministic applications, and do not account for the ability of foundational models to be executed non-deterministically \cite{bandara2026ai}. Organizations now have to deal with several compliance requirements simultaneously and need a single machine-readable evidence format.

\subsection{Operational Gaps and Runtime Evidence Challenges}

The challenge lies in translating such high-level regulations into software development practices \cite{jarvers2026engaged}. The methodologies used to ensure compliance are generally static in nature \cite{morgan2026ai}. However, the static approach does not work for an adaptive systems like AI agents. Due to the complex failure patterns of fundamental models and multi-agent workflow processes, it cannot be assured, through ex-ante benchmarking, that the alignment will persist post-deployment \cite{morgan2026ai, seth2025bridging}. There is a structural governance problem in this scenario. One cannot govern what one does not see, and static approaches do not measure the runtime execution states, tooling failure or prompt injections \cite{bandara2026ai}.

This gap underpins the importance of contemporary governance that should be based on continuous runtime compliance and assurance based on telemetry \cite{bandara2026ai, ugarte2026making}. To do so, governance should involve transformation of technical execution telemetry into machine-verifiable evidence, for example, standardized Open Security Controls Assessment Language (OSCAL), right in the process of its execution \cite{ugarte2026making}. In this respect, XAI metrics can be used as governance primitives to serve as verifiable indicators of model behavioural integrity and prevent alignment faking \cite{seth2025bridging}. Continuous verification of runtime behaviour through collection of physical, system, and application-level telemetry allows an organization to continuously monitor compliance with various regulatory frameworks \cite{bandara2026ai, ugarte2026making}. To achieve this goal, there should be a coherent governance control stack which would be able to bind execution telemetry to organizational policies \cite{morgan2026ai}. Overall, the necessity to transform real-time telemetry into machine-verifiable compliance evidence motivates the need for OTel infrastructure operating directly in the execution path. This requirement is represented by a natural transition to the architecture of Traccia, which is detailed in the next section.

\section{Architecture of Traccia}

The architecture of Traccia is composed of a continuous seven-layer pipeline of telemetry and governance designed to connect the execution of models to the administrative compliance needs of enterprises. Instead of considering observability and compliance as isolated and afterthought processes, this vertical pipeline includes the collection of telemetry as an intrinsic step of the runtime process of the application \cite{bandara2026ai, ugarte2026making}. The layers are organized in a hierarchical manner creating a data-flow pipeline where the unstructured execution events get intercepted, structured, enriched, visualized, analysed, audited, and then compiled to machine verifiable compliance evidence \cite{jarvers2026engaged}. With the help of the OTel standard as the telemetry fabric, the framework ensures that the low-level computation operations are causally connected with the high-level assertions of compliance \cite{rahman2026detecting, spadacini2026clinical}. The following subsections explain each layer of this pipeline starting from the client-side execution environment and going through the cryptographic compliance evidence exportation as illustrated in Fig. 1.

\subsection{AI Systems}

The AI systems is the boundary layer of the external operational system, where the workloads creating the telemetry stream are running. The execution of LLMs, agentic workflows, multi-agent systems, and augmented tooling pipelines created on top of orchestrators \cite{wawer2025integrating, yao2025a}. By processing user queries and handling them, the application creates a dynamic reasoning loop, querying vector database, and doing external API calls. These interactions become the unprocessed inputs for the Traccia. While the logic of the application is implemented solely in this layer, it is actively intercepted using the SDK. The execution results in this layer becomes the operational events, including model invocation and tool execution, which are passed further to the instrumentation layer. Thus, governance framework becomes an observer, not affecting the performance of the application in any way, but collecting all the information needed to create a comprehensive audit trail \cite{higeartaigh2020overcoming, kurtz2026who}.

\subsection{Runtime Instrumentation}

This layer acts as the client-side ingestion boundary that takes care of intercepting the execution flow of AI systems workflows, and translates the same into structured OTel primitives. The conventional logging approach is inadequate at handling the inherent recursive and non-deterministic nature of modern agentic execution workflows. However, Traccia incorporates a runtime instrumentation mechanism that effectively captures execution in-process. Based on the OTel framework in its native form, this layer translates the raw execution data into OTel spans, contexts, and attributes \cite{rahman2026detecting, spadacini2026clinical}. Through start end times, and execution status information, this layer creates the basic telemetry used by all subsequent layers for tracing execution path of the application. 

\begin{figure}[htbp]
  \centering
  \includegraphics[width=0.9\textwidth]{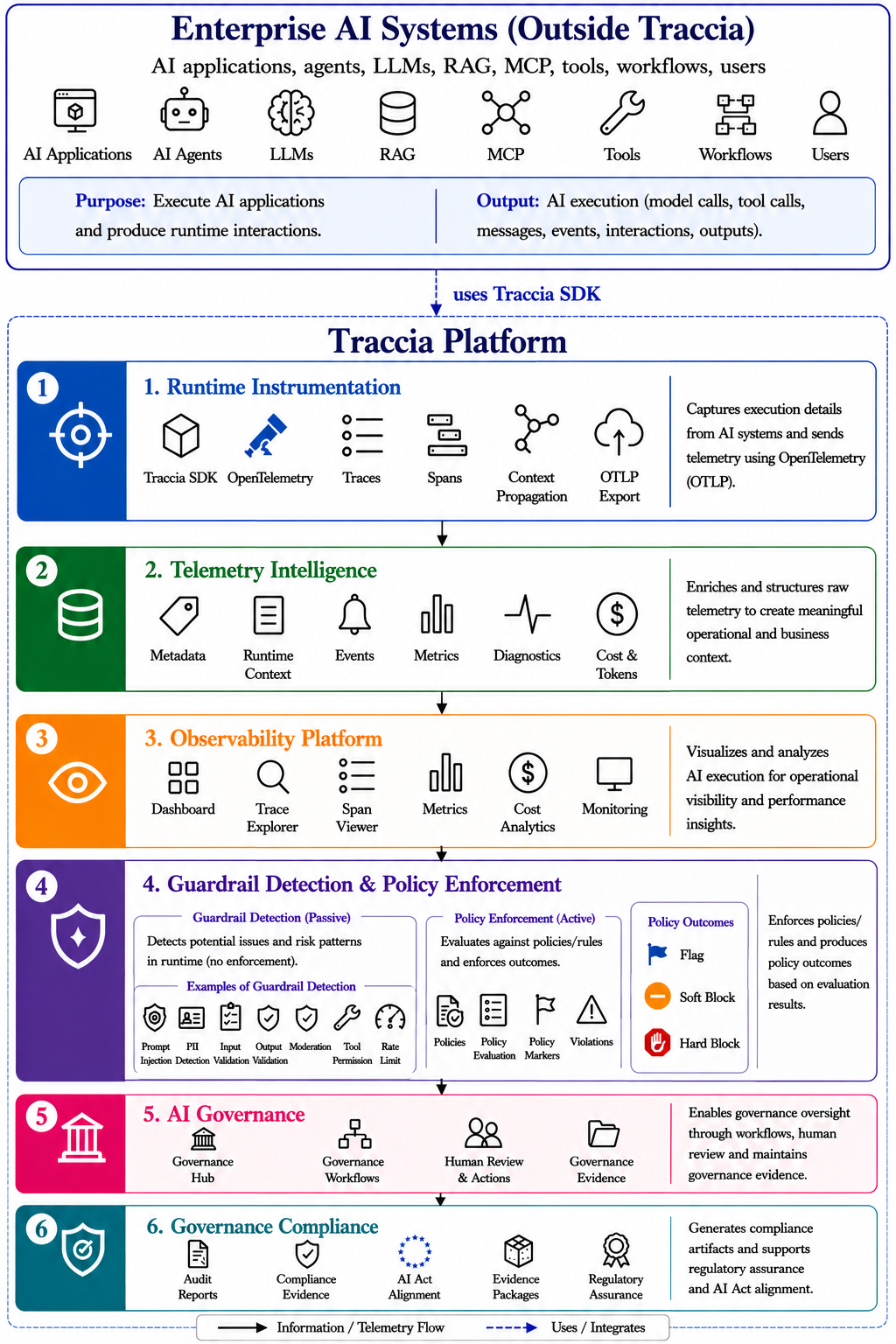}
  \caption{The vertical architecture of Traccia showing the end-to-end transformation of AI application execution into governance and compliance evidence.}
  \label{fig:arch}
\end{figure}

This instrumentation can be programmatically defined by calling the init() function, which sets up the global tracer provider, sets up OTLP/HTTP exporter settings, and sets up default sampling rates. The SDK also offers automatic patches for well-known providers and orchestrators (e. g., OpenAI and Anthropic) based on monkey-patching and event listeners. This allows automated tracking of token counts and model parameters. In order to not to pollute the source code, @observe() decorator is used for agent methods, which allows automatic span generation and logging of input/output parameters. The span() method can be used for manual span generation and high-cardinality tagging by using span.set\_attribute(). The default max\_span\_depth (= 10) limits number of spans nested within one another, to prevent possible recursive bugs leading to system memory exhaustion. Traces propagation across network boundaries is performed via W3C Trace Context headers, which ensures that the parent-child span relation is preserved in microservices with multiple agents \cite{patra2026safetyconstrained, yao2025a}. At this boundary, the SDK provides the ability to mask sensitive data using regex-based pattern matching by using init(redact\_pii=True) and redacting PII, before serializing and exporting telemetry as OTLP json lines to the platform endpoint \cite{mkander2023auditing}.

\subsection{Telemetry Intelligence}

The telemetry intelligence layer functions as the backend processing engine, taking OTLP trace payloads from the runtime instrumentation layer and enriching them into domain-specific operational intelligence. While OTel traces are excellent for tracing the distributed execution flow, they do not include the necessary machine learning-specific metadata and ontology for AI governance \cite{rahman2026detecting, spadacini2026clinical}. Processing of such traces includes parsing of span attributes and creation of a DAG to represent the full multi-agent execution hierarchy. The layer extracts and normalizes token usage, latency statistics, and names of the models used for inference across various LLM vendors \cite{chen2021evaluating, demirhan2026llmbased}. Moreover, there is a local pricing resolution engine within the layer that computes real-time execution cost estimates using local model rates or price tables provided via environment variables (for instance, agent\_dashboard\_pricing\_json). The resulting output of this layer is an enhanced semantic execution graph and structured system metrics, which are then stored in the databases of the platform and become input to the subsequent visualization and policy evaluation \cite{patra2026safetyconstrained, yao2025a}.

\subsection{Observability Platform}

The observability platform layer transforms the enhanced telemetry data and execution DAGs produced by the telemetry intelligence layer into interactive visualizations and operational alerts. This layer offers a set of analytics views that include the Dashboard, Trace Explorer, Span Viewer, and the cost allocation panels. These perspectives are used by systems engineers and compliance professionals to analyse the time delay in execution, establish the parent-child relationship, and conduct cost allocation according to agents, groups, and projects \cite{alami2026artificial, joshi2026a}. Declarative policies (set under the settings) are executed at ingestion time to assess whether the incoming metrics meet a certain threshold level (Spend Cap, Retry Protection, Duration Limit, Token Limit, Error Rate). If a metric violates the policy’s threshold, the platform logs a policy violation and creates an alert. The output of this layer is a combination of operational views and real-time alerts, which ensure the necessary transparency in the audit process \cite{masonwilliams2025reproducibility}.

\subsection{Guardrail Detection and Policy Enforcement}

This layer involves evaluating the runtime telemetry based on safety and alignment policies. Unlike metrics-based policies in the observability platform, this layer works as an OTel span processor that actively monitors completed spans for safety signals in three different confidence levels, which include Tier A (annotated spans either by guardrail\_span() or @observe(as\_type=‘guardrail’))), Tier B (provider specific signals including finish\_reason= ‘content\_filter’ or safety rating), and Tier C (heuristic keywords matching of tool error messages) \cite{hazime2025evaluation, seth2025bridging}. In this case, the engine detects any policy violations in important categories like prompt injection, PII exposure, input/output validation failures, and rate limit violations. According to the configuration of the rules/policies, the engine assigns a span with one of the following results, namely log\_only (flag), warn (soft block), and block (hard block). After the completion of root span, a missing-guardrail evaluator infers the agent capabilities (calls\_llm, uses\_tools) and identifies missing guardrails in a guardrail.summary attribute and gives developers an opportunity to find the components not covered by any safety measures \cite{collins2026systemic, nejadgholi2025a, westover2026when}.

\subsection{AI Governance}

The AI governance layer represents the administrative process of formalization of the runtime evidence and guardrails’ outputs generated by previous layers. This layer is built around the Governance Hub that serves as the enterprise-wide registry and a decoupled management tool for workspaces \cite{tabassi2023artificial, ugarte2026making}. Within the Governance Hub, compliance officers enroll agents within the AI Systems Registry by providing details of their commercial objectives, relationships with third-party vendors, and risk profile (low, medium, or high). The human-in-the-loop queues are created within this layer in order to have the compliance officers review flagged trace IDs and documenting their approval or rejection together with operational telemetry data \cite{minkkinen2022what, mkander2023auditing}. Furthermore, there is an Incident Management system inside this layer, which logs the incidents that happen during operations and classifies according to their severity-rates. The AI governance layer connects the automatic runtime telemetry to the manual registry and incident log entries in order to create the full socio-technical documentation. This documentation captures the decision-making process as well as all the corrective actions made by human administrators as a result of policy application \cite{higeartaigh2020overcoming, mkander2023auditing}. Such records will be preserved in the separate governance database and not influenced by the scheduled background tasks that clean up raw trace data according to the retention policies defined by the billing plan. The output of the layer is the governance records, which become the input of the compliance reporting.

\subsection{Governance Compliance}

The governance compliance layer is the last layer in the vertical stack and is concerned with exporting the governance information generated in the prior layer in a way that makes it possible to generate compliance reports as required for audits and regulation. Conventionally, compliance checks have been performed through subjective self-assessment processes which do not have any connections with reality \cite{ugarte2026making}. Traccia fixes the problem through automatic generation of compliance evidence from machine-generated telemetries and administrative reviews into standardized audit exports. However, legal conformations and database registration lie beyond the scope of the technical solution \cite{bandara2026ai}. Important aspects of this layer include Audit Reports, Compliance Evidence, and Evidence Packages. The platform consolidates policy configuration, run history, incident reporting, and human supervision information into cohesive Evidence Packages, which are then exported as a single JSON file protected by a cryptographic integrity hash. In order to support organizations in their efforts to comply with the EU AI Act (Articles 12, 14, 19, 26(6), 50, 72, 73, and Annexure VIII), the platform provides an optional compliance package \cite{larrauri2026reclaiming, novelli2023accountability}. Once enabled, this feature enriches exported telemetry with eu\_ai\_act\_article\_mapping metadata that links each evidence record to the relevant provisions of the EU AI Act. For example, telemetry may be associated with requirements for logging (Articles 12 and 19/26(6)), human oversight (Article 14), transparency obligations (Article 50), or incident reporting (Articles 72 and 73), thereby simplifying compliance audits and evidence retrieval. The compliance package is also able to generate pre-populated JSON structures ready for manual upload to the EU AI database (Annexure VIII). Moreover, it provides templates and outlines for compliance documentation, including technical documentation and instructions for use, reducing the administrative effort required to prepare regulatory submissions. As a result, compliance packages based on trace-based execution lineage provide a tamperproof chain of compliance documentation that meets the requirements of both enterprise risk management and international regulation \cite{chua2026shadow, tripathi2026mapping}.

\subsection{Implementation Example of Traccia}

In this section, an example is given of the full implementation lifecycle of a high-risk deployment and how Traccia translates it into a governable and evidentiary system. Instead of listing the functionality of the SDK, we go through the end-to-end lifecycle of a specific, illustrative, high-risk deployment of an application (an AI Loan Approval Assistant) from runtime instrumentation and inline governance enforcement, through model inference and governance enrichment to human-in-the-loop review and compliance evidence creation. The idea is to show that the governance responsibilities placed on high-risk AI applications under the EU AI Act (such as logging, transparency, human oversight, incident management, and fundamental rights assessment) can be achieved as a natural side effect of running the application, without adding any special compliance code to the application itself. The loan approval was specifically used as the scenario since it clearly falls into Annex III of the EU AI Act, where automated creditworthiness assessment has a significant impact on the ability of an individual to obtain financial services and hence triggers all governance responsibilities. Figure 2 depicts the visualization of a governed execution trace provided by the platform. The important aspect shown in Table 1, where phases 1-3 are locally and inline implemented, without the telemetry leaving the host; phase 4 works on the exported telemetry and generates the documentation for regulators. Both phases are linked by one consistent trace of evidence.

\begin{table}[htbp]
\centering
\caption{The lifecycle comprises four phases, each mapped to a concrete artifact produced at runtime.}
\label{tab:lifecycle}
\begin{tabularx}{\textwidth}{lXX}
\toprule
\textbf{Phase} & \textbf{Runtime Component} & \textbf{Governance Artifact Produced} \\
\midrule
Instrumentation & \texttt{init()} and OpenTelemetry bootstrap & Agent registration, risk-tier declaration, runtime trace context \\
Guardrail Enforcement & Prompt injection and PII guardrails & Guardrail findings, enforcement decisions, redaction records \\
Inference and Enrichment & Auto-instrumented LLM invocation + attributes for governance log enrichment & Model metadata, latency, token usage, integrity hashes, transparency markers \\
Governance and verification & Governance Hub & Human reviews, incident records, FRIA, evidence bundles, audit log \\
\bottomrule
\end{tabularx}
\end{table}

\begin{figure}[htbp]
  \centering
  \includegraphics[width=0.9\textwidth]{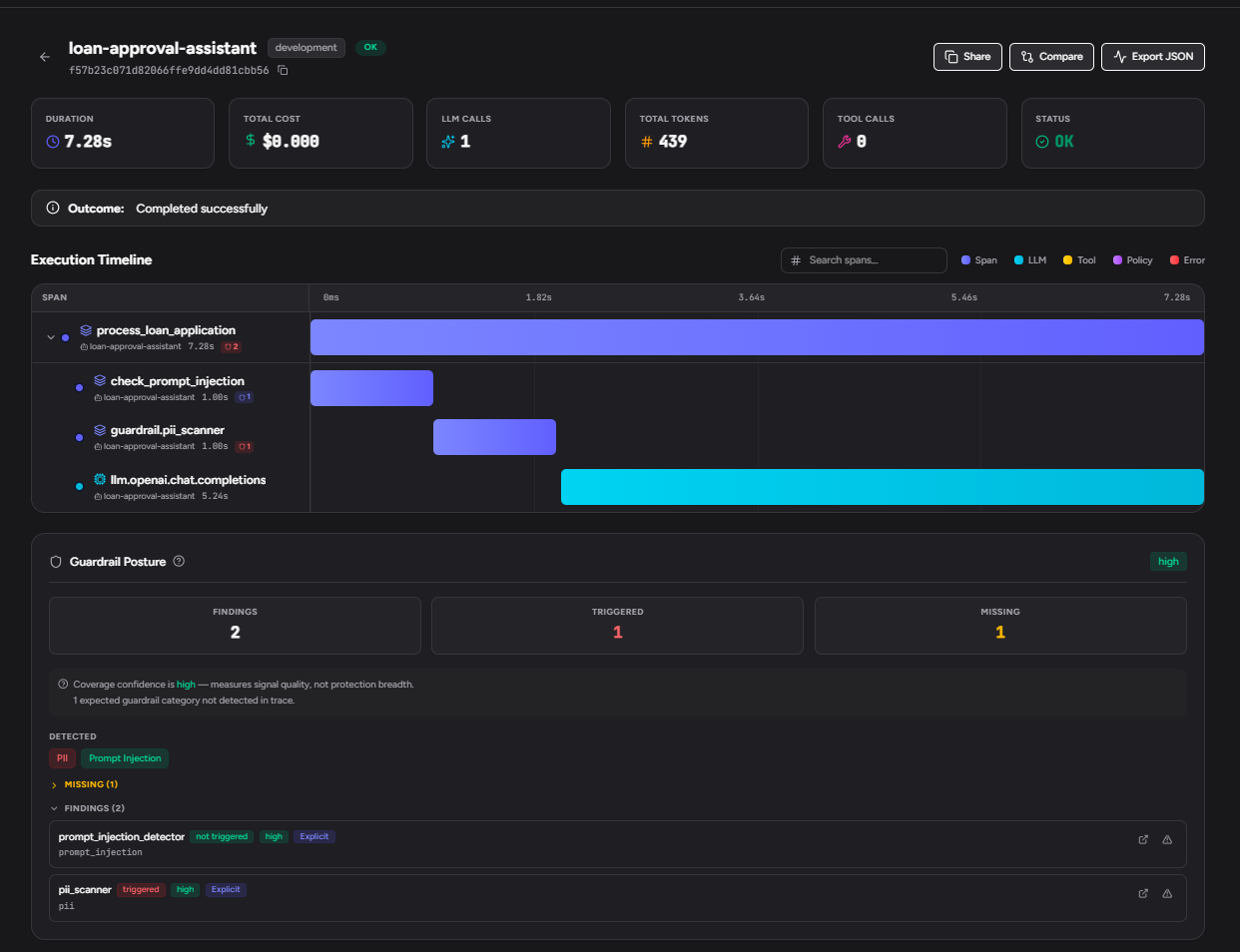}
  \caption{Execution trace from Traccia platform for the AI Loan Approval Assistant (trace id - f57b23c071d82066ffe9dd4dd81cbb56).}
  \label{fig:trace}
\end{figure}

The application initializes its governance context when started using a single initialization call. This process involves registering the agent, declaring its governance class, and enabling the local processing pipeline (agent enrichment, guardrail extraction, PII redaction, and OTLP export):

\begin{lstlisting}[language=Python]
init(
api_key=API_KEY,               # loaded from environment; not hard-coded
    	agent_id="loan-approval-assistant",
    	compliance={
        		"frameworks": ["eu_ai_act"],
        		"risk_tier": "high",
    		},
    	redact_pii=True,
    	enable_console_exporter=True,
)
\end{lstlisting}

The compliance schema plays an important role: through setting the value of “risk\_tier = high” at the time of initialization, the deployer sets the regulatory starting point against which every span will be assessed thereafter. Thereafter, every operation performed by the application will become part of a single nested execution trace, with every span carrying forward the identity, scope, and risk level of the agent. The fulfilment of the condition precedent for Article 12 (record-keeping) is thus ensured through logging.

Before any model inference occurs, two guardrails are evaluated against the incoming request. The design principle is fail-closed enforcement at the boundary: unsafe requests are rejected before the LLM is ever reached. The first guardrail, declared via an explicit decorator, detects prompt-injection and jailbreak attempts and operates in block mode:

\begin{lstlisting}[language=Python]
@observe(
as_type="guardrail",
    	attributes={
        		"guardrail.name": "prompt_injection_detector",
        		 "guardrail.category": "prompt_injection",
        		"guardrail.enforcement_mode": "block",
        		"guardrail.policy_id": "policy-loan-inj-v1.0",
    },
)
\end{lstlisting}

In the case of a detected manipulation (e.g., “disregard prior instructions”), the request is interrupted, and the control logic never makes it to the underlying model. It is crucial that this response is not made in silence, as the guardrail interval contains information about the trigger, its enforcement method, and the identifier of the controlling policy.

The second guardrail identifies any personally identifiable information and runs in the warn mode in a context-managed span:

\begin{lstlisting}[language=Python]
with guardrail_span(
	 "pii_scanner", 
	category="pii",
	enforcement_mode="warn") as gs:
	...
\end{lstlisting}

In the case of detection of sensitive tokens like email address, the text provided by the user gets redacted using the function redact\_string(). Thus, when the payload is sent to the model as well as when telemetry is exported from the system, redacted strings (like [REDACTED\_EMAIL]) are included, rather than PII. The data minimization principles of the Act are met in this way.

Once both guardrails have been passed, the inference request is issued by the application. The model is called through the OpenRouter API endpoint via an OpenAI compatible client. Since Traccia automatically instruments the client, no further development effort is needed to instrument the call:

\begin{lstlisting}[language=Python]
response = client.chat.completions.create(
    	model=model_name,
    	messages=[...],
    	temperature=0.2,
)
\end{lstlisting}

Execution latency, the resolved model ID, and tokens consumed by the prompt, the completion, and the total are captured automatically in the llm.openai.chat.completions Span that gets generated. These two engineering problems are solved at once through performance observability and localized cost tracking, none of which is possible through typical application metrics. When the recommendation is generated, governance metadata is added to the parent span:

\begin{lstlisting}[language=Python]
gov_attrs = enrich_governance_attributes(
   	event_type="inference",
   	model_id=model_name,
   	input_text=prompt,
   	output_text=decision,
    	session_id="session-loan-2026",
    	eu_risk_tier="high",
)
\end{lstlisting}

Enrichment adds additional metadata about the governance-related nature of the AI event, such as type of AI event, model identifier, session information, cryptographic integrity hashes for both the input and output, EU AI Act risk classification, and transparency/disclosure information. Cryptographic integrity hashes are an essential part of the evidence model since they bind the decisions to their input/output in a way that makes any modification evident. It guarantees that the persisted evidence is a faithful representation of the initial interaction. Before sending telemetry out of the system, the redaction process takes place to exclude/replace any sensitive information from the persisted trace. The above steps create runtime telemetry, which this step transforms into verifiable governance evidence via the Governance Hub. Steps include: registration of the assistant as high-risk AI system makes all subsequent governance actions to be tied to specific deployment and version. The traces will then be reviewed through the review process, which captures decision of approval, identity of the reviewers and any comments, operationalizing the requirement of human oversight in Article 14 by creating a link between an automated decision and the human who approved it. Detected prompt injection attacks and lack of complete guardrails coverage (through gap analysis by the MissingGuardrailEvaluator) will be reported as governance incidents and managed through an explicit lifecycle from open to resolved, thereby creating an incident trail as per Articles 72-73. Completion of the FRIA through the seven-step wizard that captures the purpose of the system, population affected, foreseeable risks, mitigation, and oversight mechanism as required in Article 27. Traccia collects evidence packages and maintains an immutable administrative audit log of the system registration, FRIA completion, review decisions, incident management, and evidence export.

In summary, the four stages show a continuing evolution process: raw telemetry of runtime evolves into structured compliance data, without any gap and without any compliance code written by developers. This example provides support for the main thesis of the paper, that runtime governance can be extracted from execution itself, and that the compliance data generated in the process can be directly mapped to real obligations of the EU AI Act (automatic logging, transparency notification, human supervision, incident management, monitoring, and FRIA preparation). It is important to note that Traccia is not a replacement of the compliance processes; on the contrary, it provides the runtime substrate (verifiable logs, guardrails, and audit trails), which make the compliance processes possible. The complete example and the associated artefacts produced can be found at \url{https://github.com/naiknutan97/traccia-paper-examples}.

\section{Comparison with Existing Platforms}

The landscape of observability and governance of AI products has become more sophisticated over the last few years with the advent of various platforms designed to manage different facets of the process of overseeing LLM deployments. In order to contextualize Traccia in this landscape, this section presents a structured comparative analysis of eight representative platforms: Arize AI (Phoenix), Langfuse, LangChain (LangSmith), Braintrust, Maxim AI, Datadog LLM Observability, Credo AI, and Holistic AI. Together, the platforms cover all major areas in which the tools can be classified – evaluation-focused engineering platforms, enterprise GRC platforms, and generic APM extensions – and therefore present a suitable base for comparison. The comparison is based on four functional dimensions that are relevant for the governance mission of Traccia: (i) telemetry architecture and vendor agnosticism, (ii) guardrails validation at runtime, (iii) compliance with regulations, and (iv) cost tracking and FinOps.

\subsection{Telemetry Architecture and Vendor Independence}

There is an important architectural difference between Traccia and most of the other systems reviewed here. The proprietary instrumentation layers (custom SDKs, agent daemons, or span processors) of Arize AI, LangSmith, Braintrust, Maxim AI, and Datadog LLM Observability ensure the connection between the telemetry pipeline of the application and the backend of a particular vendor. Langfuse uses its own protocol with OTel serving as a secondary ingestion channel. Traccia, however, is designed as a pure OTel sidecar, producing generic W3C OTLP spans that can be ingested by any compatible collector, such as Jaeger, Grafana, and Prometheus, without changing the code \cite{bandara2026ai, morgan2026ai}.In this regard, it is clear that the system offers some practical benefits since leaving the proprietary platform would require changes to the instrumentation code, while switching from Traccia would only require pointing another OTLP exporter endpoint. What is more, Traccia provides provider auto-instrumentation through monkey-patching on the level of imports without any specific integration code and modifications of application's logic. Enterprise GRC platforms (Credo AI, Holistic AI) do not perform any telemetry instrumentation during code execution and operate at a higher organizational policy layer \cite{eisenberg2025the}.

\begin{table}[htbp]
\centering
\caption{Telemetry architecture comparison across platforms.}
\label{tab:telemetry_comp}
\begin{tabularx}{\textwidth}{lXXXX}
\toprule
\textbf{Platform} & \textbf{Tracing Standard} & \textbf{Agent/Daemon Required} & \textbf{Auto-Instrumentation} & \textbf{Vendor Lock-in} \\
\midrule
Traccia & Pure OTLP (OTel) & No & Yes (import-level) & None \\
Arize / Phoenix & OTel + OpenInference & No & Partial & Medium \\
Langfuse & Custom + OTel bridge & No & Drop-in wrappers & Medium \\
LangSmith & Proprietary & No & Decorator-based & High \\
Braintrust & Custom (OTel processor) & No & SDK-driven & High \\
Maxim AI & Proprietary + OTLP ingest & No & SDK-driven & High \\
Datadog & ddtrace (proprietary) & Yes & Decorator-based & Very High \\
Credo AI & N/A (policy layer) & No & N/A & N/A \\
Holistic AI & N/A (offline audit) & No & N/A & N/A \\
\bottomrule
\end{tabularx}
\end{table}

\subsection{Runtime Guardrail Verification}

Runtime guardrail verification, defined as the detection and classification of active safety mechanisms on real LLM spans, is the particular aspect where Traccia clearly sets itself apart from the other products that already exist. None of the listed platforms feature a native tool for verifying the presence of guardrails and their runtime activation. The evaluation-oriented platforms (Arize AI, Langfuse, LangSmith, Braintrust, Maxim AI) include quality-oriented evaluators in the form of toxicity assessment, hallucinations detection, or faithfulness assessment, but these tools check the properties of the output, not the presence of a safety wrapper. Datadog implements a custom processor hook to redact sensitive data, thus covering only one narrow part of the spectrum of safety considerations. Credo AI and Holistic AI manage policies for guardrails via a self-reporting questionnaire and risk assessment system, resulting in documentation of the intended safety approach, not the actual runtime enforcement of guardrails \cite{eisenberg2025the}.

Traccia resolves this problem via a three-layer guardrail classification system. Layer A detects explicit guardrails defined by developers and discernible in the trace structure. Layer B detects native safety systems in the provider's system (such as OpenAI’s content moderation flags). Layer C uses heuristic pattern detection for spans where there is no explicit guardrail definition. Moreover, Traccia comes equipped with a missing guardrail evaluator, which alerts LLM spans that run without any safety wrapper. This feature proves especially relevant to the new European Union (EU) AI Act as the law mandates that providers of high-risk AI systems develop technical safety measures and provide documentation of said measures, which the trace-level evidence helps fulfil \cite{kim2025ai, ugarte2026making}.

\subsection{Regulatory Compliance Support}

Under the regulatory compliance category, there is a definite division among the platform vendors. Engineering-based platforms such as Arize AI, Langfuse, LangSmith, Braintrust, Maxim AI, and Datadog offer no capabilities for mapping application telemetry into regulatory frameworks. These platforms’ compliance offerings are restricted to the infrastructure certification of the platforms (SOC 2, ISO 27001) and not any tools which could help the customer show compliance for the AI systems they run. In contrast, the two enterprise GRC platforms, namely Credo AI and Holistic AI, cover all aspects of the regulatory frameworks. For instance, Credo AI offers policy packs according to NIST AI RMF, EU AI Act, ISO 42001, and SOC 2. Holistic AI also has policies and risk categorizations at the framework level, similar to the aforementioned standards \cite{eisenberg2025the}. But interestingly, both platforms provide compliance evidence through self-assessments and audits.

Traccia holds an exceptional place in this domain since it provides compliance evidence through application telemetry. Traccia's EU AI Act component correlates OTel span attributes with particular articles of Regulation (EU) 2024/1689 using a Governance Enrichment Processor, adding regulatory tags to the trace at runtime. Evidence Packs that include JSON audits of traces labelled with guardrail status, costs, and articles are an example of regulatory artifacts providing regulator and auditor evidence of system behaviour and not policies \cite{morgan2026ai, ugarte2026making}. A wizard for FRIA is designed to help fulfil the requirements of Article 27, relevant to operators of high-risk AI systems \cite{kim2025ai, ugarte2026making}. Other regulatory domains (GDPR, HIPAA, SOC 2) are currently planned in the product development roadmap.

\begin{table}[htbp]
\centering
\caption{Regulatory compliance capabilities across platforms.}
\label{tab:compliance_comp}
\begin{tabularx}{\textwidth}{lXXXX}
\toprule
\textbf{Platform} & \textbf{EU AI Act Module} & \textbf{Evidence from Telemetry} & \textbf{FRIA Support} & \textbf{Governance Hub} \\
\midrule
Traccia & Yes (article mapping + evidence packs) & Yes & Yes (Art. 27 wizard) & Yes \\
Arize / Phoenix & No & No & No & No \\
Langfuse & No & No & No & No \\
LangSmith & No & No & No & No \\
Braintrust & No & No & No & No \\
Maxim AI & No & No & No & No \\
Datadog & No & No & No & No \\
Credo AI & Policy packs (self-reported) & No & No & Yes (GRC dashboards) \\
Holistic AI & Framework mapping (self-reported) & No & Risk scoring templates & Yes (AI inventory) \\
\bottomrule
\end{tabularx}
\end{table}

\subsection{Cost Tracking and FinOps}

LLM inference workloads have a unique observability problem as far as cost attribution is concerned due to changing provider pricing policies. In some cases, API endpoints use token usage and generation costs as pricing metrics, whereas others use elapsed time or request counts, while self-hosting involves compute costs allocation. Prompt caching and batch pricing, in addition, make it harder to keep track of costs in real time. At the platform level, there is an engine that performs retroactive recalculations of trace costs based on a database of more than 2,500 model pricing configurations updated continually, enabling organizations to amend historical financial telemetry with changes in provider pricing and tiering policies. This makes sure that the records of costs stay up-to-date, which is important in terms of the companies' financial reporting and FinOps practices \cite{monfared2026timing, rahman2026detecting}. Neither Credo AI nor Holistic AI support cost tracking since their functionality is confined by organizational and algorithmic layers respectively.

\subsection{Evaluation, Prompt Engineering, and Quality Assurance}

However, in addition to runtime governance and compliance, there are significant differences between the surveyed platforms regarding the maturity level of evaluation, prompt engineering, and quality assurance. These processes facilitate the LLM development process, allowing developers to create prompts, assess the behaviour of models, and detect any regressions and increase the quality of applications being developed and deployed. Prompt management, which includes prompt versioning, prompt sandboxing, prompt comparison, and rollback, is provided natively by Langfuse, Langsmith, Arize AI, and Braintrust using versioned prompt registries. Arize AI offers Prompt Hub within its Phoenix experimentation platform, and Maxim AI has a multi-provider prompt playground. Datadog and Holistic AI lack prompt engineering environments.

Automated evaluation has emerged as an integral part of the contemporary process of LLM engineering due to the LLM-as-Judge approach \cite{chen2021evaluating, mkander2023auditing}. Langfuse, LangSmith, Arize AI, Datadog, Braintrust, and Maxim AI provide configurable evaluation pipelines with different levels of customizability. LangSmith and Arize AI allow for customizable judge prompts and experiment management, Datadog provides out-of-the-box evaluators for hallucination, toxicity, and faithfulness, along with evaluation classes that can be extended, while Braintrust enables trial-based evaluation and immutable experiment snapshots for regression analysis. Maxim AI boasts of the widest range of pre-built evaluators with more than thirty AI and statistical metrics available. LangSmith and Datadog allow for online evaluations of production traffic, while Langfuse and Maxim AI provide some level of support for it. Evaluation functionality will be provided in the future versions of Traccia. Pre-built evaluation metrics vary greatly in their number. Faithfulness, hallucination detection, toxicity, RAG evaluation metrics (context precision, context recall, answer relevance), and agent trajectory evaluation metrics are widely supported. One of the special features of Maxim AI is that it offers classical NLP metrics like BLEU, ROUGE, and F1 Score as well as voice metrics like Word Error Rate (WER) and Signal-to-Noise Ratio (SNR). Unlike Maxim AI, Holistic AI works on pre-deployment statistical bias, fairness, robustness, and explainability metrics of traditional ML models, not prompt-level metrics \cite{alsayed2026operational}.

Human-in-the-loop reviewing functions are available with Langfuse, LangSmith, Arize AI, Datadog, and Maxim AI via annotation queues, labelling interface, and dataset export capability. Braintrust offers only partial annotation functionality while Holistic AI uses audit datasets off-line. Human annotation pipelines are on the roadmap of Traccia. The differences among the surveyed platforms are also evident in CI/CD integrations, developer tooling, and debugging support. Braintrust offers most comprehensive developer tooling via bt eval CLI, multi-language SDKs, and experiments management, while LangSmith and Datadog offer GitHub Actions based evaluation integration. Self-hosted deployments are possible with Langfuse, LangSmith, and Arize AI. The most advanced functionality related to AI-assisted debugging is offered with Arize AI via Alyx assistant, while LangSmith and Braintrust offer structured regression analysis and root cause comparison tools. These features are out of scope of Traccia at this point, which aims to be a runtime governance, guardrail validation, and telemetry-based compliance solution to complement evaluation-focused AI engineering platforms.

\subsection{Positioning and Complementarity}

The comparative analysis presented in this section demonstrates that the current AI tooling ecosystem is characterized by three complementary categories of platforms, each addressing a different phase of the AI system lifecycle. The first category comprises evaluation-oriented engineering platforms, including Langfuse, LangSmith, Arize Phoenix, Braintrust, Maxim AI, and Datadog LLM Observability. These platforms primarily support prompt engineering, experimentation, automated evaluation, human annotation, regression testing, and application debugging, enabling developers to iteratively improve model quality before and after deployment. Their primary objective is to optimize model behaviour, application performance, and developer productivity rather than to provide governance evidence. The second category consists of enterprise governance platforms, represented by Credo AI and Holistic AI. These systems operate primarily at the organizational level by maintaining AI inventories, managing governance policies, performing risk assessments, supporting regulatory documentation, and coordinating compliance workflows. However, their focus lies on governance processes at an organization-wide level as well as on regulatory management processes, but not on the observation of runtime system behaviour. Thus, the evidence for compliance comes from manual maintenance of documents and questionnaires and not from telemetry. The suggested framework can be regarded as the third type of framework, which might be called runtime governance infrastructure. In contrast to evaluation frameworks, it does not offer any attempts to create a complete environment for prompt engineering, benchmark management, or quality evaluation pipeline although some of these elements will appear in future versions. In the same way, it does not substitute enterprise governance frameworks that manage organizational policies and corporate compliance processes. The suggested framework works directly in the path of the application of AI and converts OTel traces into governance-enabled telemetry containing guardrail metadata, regulatory metadata, costs, and compliance evidence \cite{bandara2026ai, morgan2026ai, ugarte2026making}.

Such a design allows for capabilities which are not available in current platforms, namely runtime validation of guardrails in production, automatic detection of the lack of safety mechanisms, continuous generation of machine-verifiable evidence of compliance, and linking of execution records to regulatory requirements such as those set out in the EU AI Act. By producing governance artifacts based on telemetry data instead of static documentation, the framework ensures an ongoing and traceable evidence trail throughout the operational lifetime of AI systems \cite{bandara2026ai, morgan2026ai, ugarte2026making}. Based on these remarks, it is evident that the proposed framework should be seen more as complementing than competing with existing platforms. While evaluation platforms are still appropriate for prompt tuning, benchmarking, regression testing, and gradual improvement of models' quality, enterprise governance tools continue to excel in the areas of organizational oversight, policy and regulation management. Therefore, the organization will be able to integrate the current evaluation platform along with the governance platform and the proposed model to develop the complete AI lifecycle from development, deployment, to runtime monitoring, and regulatory compliance \cite{ugarte2026making}.

\section{Summary and Future Work}

The current article has introduced Traccia, a unique telemetry and governance pipeline that enables enterprise-level AI systems, such as the execution of LLMs and autonomous AI agents for compliance and governance. Building on the OTel standards for capturing the semantic attributes of AI-related data flows, Traccia consistently translates raw execution streams into structured, auditable compliance evidence to reduce the risks of SaaS integration and shadow AI development. The vertical stack starts at the level of instrumentation and token-level costs, then proceeds to back-end telemetry intelligence, and ends with passive guardrails and Governance Hub. The integration of the above-mentioned layers allows for the automation of the collection of tamper-proof telemetry logs, traceable reviews by human operators, and incident logs. Most importantly, the opt-in compliance module allows correlating the operational records collected by Traccia with the legal obligations set by the EU AI Act. Future work will involve developing Traccia further in many critical aspects. First, whereas the present system is mostly in an ‘observe and alert’ mode of operation, future systems will explore the possibility of having runtime enforcement capabilities, through which policies can actively block the operations of agents. Second, adapters will be developed to support compliance with other international frameworks such as the HIPPA, SOC 2, ISO/IEC 42001 standard, and US executive orders.

\section*{Declaration of generative AI and AI-assisted technologies in the writing process}

During the writing of this paper, the author(s) have made use of OpenAI’s ChatGPT for grammar corrections. Following this, the author(s) have made necessary edits to the paper and are fully responsible for the content of the publication.

\section*{Declaration of competing interest}

The authors declare that they have no known competing financial interests or personal relationships that could have appeared to influence the work reported in this paper

\bibliographystyle{unsrt}
\bibliography{references}

\end{document}